\documentclass{article} 
\usepackage{collas2022_conference,times}

\usepackage{amsmath,amsfonts,bm}

\def\eqref#1{equation~\ref{#1}}

\def\1{\bm{1}}

\DeclareMathAlphabet{\mathsfit}{\encodingdefault}{\sfdefault}{m}{sl}
\SetMathAlphabet{\mathsfit}{bold}{\encodingdefault}{\sfdefault}{bx}{n}

\usepackage{graphicx}
\usepackage{varwidth}
\usepackage{amsmath}
\usepackage{amssymb}
\usepackage{booktabs}
\usepackage{algorithm}
\usepackage[noend]{algpseudocode}
\usepackage{multirow}
\usepackage[T1]{fontenc}
\usepackage{bbm}

\usepackage{hyperref}
\hypersetup{
    colorlinks=true,
    linkcolor=red,
    filecolor=magenta,      
    urlcolor=blue,
    citecolor=purple,
    pdftitle={Overleaf Example},
    pdfpagemode=FullScreen,
    }

\makeatletter
\newcommand{\printfnsymbol}[1]{%
  \textsuperscript{\@fnsymbol{#1}}%
}

\title{Task Agnostic Representation Consolidation: a Self-supervised based Continual Learning Approach}

\author{Prashant Bhat, Bahram Zonooz\thanks{Shared last author.}, Elahe Arani\printfnsymbol{1}\\
Advanced Research Lab, NavInfo Europe, The Netherlands\\
\texttt{\{prashant.bhat, elahe.arani\}@navinfo.eu, bahram.zonooz@gmail.com}
}

\usepackage[capitalize]{cleveref}
\crefname{section}{Sec.}{Secs.}
\Crefname{section}{Section}{Sections}
\Crefname{table}{Table}{Tables}
\crefname{table}{Tab.}{Tabs.}

\collasfinalcopy 

\begin{document}
\maketitle

\begin{abstract}
   Continual learning (CL) over non-stationary data streams remains one of the long-standing challenges in deep neural networks (DNNs) as they are prone to catastrophic forgetting. CL models can benefit from self-supervised pre-training as it enables learning more generalizable task-agnostic features. However, the effect   of self-supervised pre-training diminishes as the length of task sequences increases. Furthermore, the domain shift between pre-training data distribution and the task distribution reduces the generalizability of the learned representations. To address these limitations, we propose Task Agnostic Representation Consolidation (TARC)\footnote{Code can be found at: \url{https://github.com/NeurAI-Lab/TARC}.}, a two-stage training paradigm for CL that intertwines task-agnostic and task-specific learning whereby self-supervised training is followed by supervised learning for each task. To further restrict the deviation from the learned representations in the self-supervised stage, we employ a task-agnostic auxiliary loss during the supervised stage. We show that our training paradigm can be easily added to memory- or regularization-based approaches and provides consistent performance gain across more challenging CL settings. We further show that it leads to more robust and well-calibrated models. 
\end{abstract}

\section{Introduction}
Computational systems that operate in the real world are exposed to the continuous stream of non-i.i.d data and are required to learn multiple tasks sequentially. 
Learning from a stream of non-i.i.d data causes the new information to overwrite the previously learned knowledge in the neural network leading to catastrophic forgetting. 
Several approaches have been proposed in the literature to address the problem of catastrophic forgetting in CL. Replay-based methods \citep{Robins1995CatastrophicFR, buzzega2020dark, ratcliff} store and replay a subset of samples belonging to previous tasks along with the current batch of samples. Regularization-based methods \citep{schwarz2018progress, si} insert a regularization term to consolidate the previous knowledge when training on new tasks. These methods avoid using memory buffer altogether alleviating the memory requirements \citep{Delange_2021}. 

Although aforementioned approaches have been partially successful in mitigating the catastrophic forgetting, they still suffer from several shortcomings.  
Since CL methods rely extensively on cross-entropy loss for classification tasks, they are prone to lack of robustness to noisy labels \citep{sukhbaatar2015training} and the possibility of poor margins \citep{poor_margin} affecting their ability to generalize across tasks. Furthermore, the optimization objective in cross-entropy loss encourages learning of representations optimal for the current task sidelining the representations that might be necessary for the future tasks, resulting in prior information loss \citep{zhang2020self}. 
Also, the representations of the observed tasks drift when new tasks appear in the incoming data stream exacerbating the backward interference \citep{representation_drift}.
Therefore, we assume task-specific learning is the root cause of these problems and is not well equipped to deal with catastrophic forgetting.

We hypothesize that learning task-agnostic representations in addition to task-specific representations can potentially mitigate aforementioned problems by improving forward facilitation while reducing backward interference in CL. Self-supervised pre-training has been widely regarded to learn task-agnostic generalizable representations \citep{MoCo, byol}.
In many real-world CL scenarios however, the data distribution of the future tasks is not known beforehand. Thus, pre-training on a different data distribution often leads to domain shift subsequently reducing the generalizability of the learned representations. Furthermore, longer task sequences diminish the effect of self-supervised pre-training as the learned representations are repeatedly overwritten to maximize the performance on the current task. Domain shift,
non-availability of pre-training data and computational costs associated with pre-training can be avoided if task-agnostic learning is integrated into CL training. 

\begin{figure*}[t]
\centering
    \includegraphics[trim=0 7 0 7 cm, width=.7\linewidth]{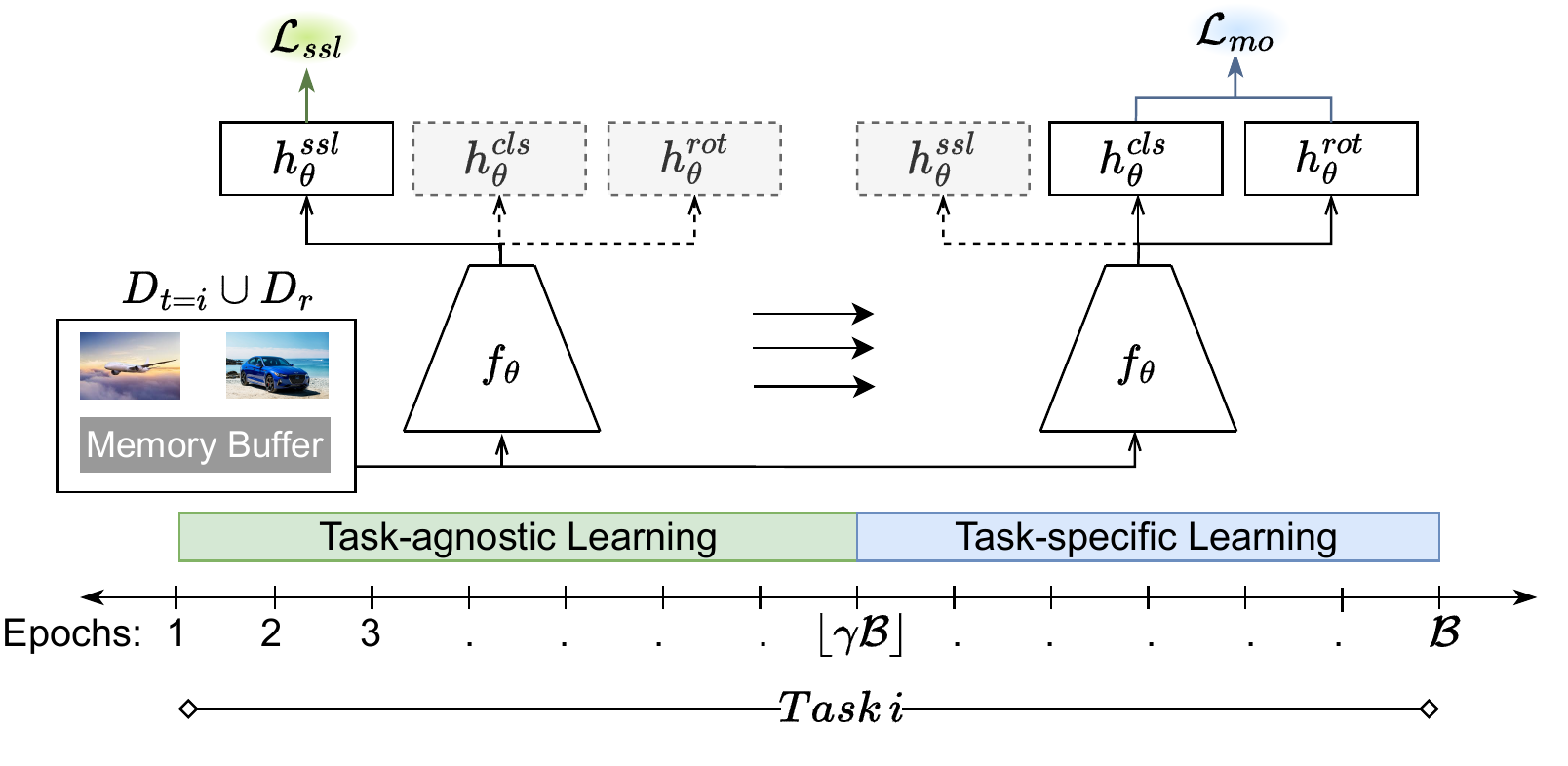}
  \caption{Proposed method. The training budget $\mathcal{B}$ for each task $i$ is divided into task-agnostic and task-specific learning phases.}
\end{figure*}

We therefore propose an online Task-Agnostic Representation Consolidation (TARC), a two-stage generic CL training paradigm that intertwines task-specific and task-agnostic learning through self-supervised learning and multi-objective learning. Our method integrates task-agnostic learning into CL training thereby avoiding problems associated with self-supervised pre-training. Our contributions are as follows: 
\begin{itemize}
    \item We propose a two-stage generic CL framework that intertwines task-agnostic and task-specific learning to train robust, well-calibrated models.     
    \item We extend our method to state-of-the-art replay-based and regularization-based methods. Our method outperforms the baseline in the more challenging CL scenarios. 
    \item We provide extensive analyses including the bias towards recent tasks and robustness to noisy labels to highlight the additional benefits our method brings without any explicit constraints. 
\end{itemize}

\section{Related Works}

\subsection{Continual learning}
Early works attempted  to  mitigate  the  effect  of  catastrophic  forgetting  by  replaying  a  subset  of  training  data from the  previous tasks stored in the replay buffer alongside samples from the new task \citep{Robins1993}. Samples in the replay buffer are used as model inputs for rehearsal and/or for constraint optimization of the new task \citep{Delange_2021,CLS-ER}. Experience Replay (ER) \citep{Ratcliff1990, Robins1995CatastrophicFR} explicitly interleaves the old samples from the replay buffer with the current batch of samples while training on the new tasks. 

Regularization-based methods on the other hand, avoid storing raw inputs, thus alleviating the memory requirements \citep{Delange_2021}. Instead, a regularization term is introduced to consolidate the previous knowledge when learning a new task. Elastic Weight Consolidation (EWC) \citep{kirkpatrick2017overcoming} estimates the importance of all parameters of a neural network and penalizes the changes to the important parameters in the later tasks. However, EWC is not scalable with large number of tasks as the number of regularization terms grow linearly with the number of tasks. Online-EWC (oEWC) \citep{schwarz2018progress} modified the original algorithm to avoid the linear growth in the computational requirements. Synaptic Intelligence (SI) \citep{si} introduced brain-inspired intelligent synapses that accumulate task relevant information over time, and exploit this information to rapidly store new memories without forgetting old ones. 

Remembering for the Right Reasons (RRR) \citep{rrr} proposed a training paradigm that additionally stores model explanations for each example and encourages explanation consistency throughout CL training. 
RRR is a generic framework that can be extended to any memory or regularization-based approaches, similar to our proposed framework. Therefore, we consider RRR as a competing approach in our analysis. 

\subsection{Self-supervised learning}
Self-supervised learning methods broadly fall into one of two categories: generative and contrastive methods \citep{liu2020selfsupervised}. Context-instance contrastive methods learn the representation of local features associative to the representation of the global context. 
More recently, InstDisc \citep{wu2018unsupervised}, MoCo \citep{MoCo} and SimCLR \citep{simclr} leverage instance discrimination as a pretext task. SimCLR learns representations by contrasting semantically similar (positive) and dissimilar (negative) pairs of data samples such that similar pairs have the maximum agreement via a contrastive loss in the latent space \citep{simclr}. However, as with metric learning, contrastive learning benefits from hard negatives \citep{cai2021are, hard_negatives}. SupContrast \citep{supcontrast} extends SimCLR to a fully supervised setting to effectively leverage the label information. It eliminates the need for hard negative mining by using several positive and negative samples per anchor sample. Also, SupContrast consistently outperforms cross-entropy loss on large scale classification problems \citep{supcontrast}. 

Each task in CL is usually constrained by a training budget, limiting model's ability to consolidate the representations for the current task. Pre-training has been traditionally used to offset the limited data/training time through transfer learning. \citet{gallardo2021selfsupervised} empirically showed that self-supervised pre-training yields representations that generalize better across tasks than supervised pre-training in CL. Owing to additional computational effort, some of the approaches, e.g. \citep{zhang2020self, representation_drift, mazumder2021fewshot, su2020does}, relinquished pre-training altogether and employed auxiliary pretext task to boost task-agnostic learning. 
However, self-supervised learning as an auxiliary loss improved the baseline only marginally.  

In this work, we adapt ER, oEWC and SI to our CL training paradigm. We employ SupContrast as a task-agnostic learning objective and rotation prediction as an auxiliary loss alongside cross-entropy loss. Our method differs from the above methods in two ways: Firstly, we do not employ any pre-training, instead achieve the same objective through task-agnostic learning during CL training. Secondly, the use of auxiliary self-supervised loss in our training paradigm is necessitated by the need to preserve task-agnostic representations. \Cref{ablation_study_section} provides a comparison of different variants of self-supervision aided CL methods.


\section{Proposed Method}
Continual learning normally consists of \textit{T} sequential tasks. During each task, the input samples and the corresponding labels $(x_{t}, y_{t})$ are drawn from the task-specific data distribution $\mathcal{D}_{t}$. Our continual learning model consists of a backbone network $f_{\theta}$ and three heads $h^{ssl}_{\theta}, h^{cls}_{\theta}$ and $h^{rot}_{\theta}$. 
These heads correspond to task-agnostic head, classification head and rotation head respectively. The continual learning model $g_{\theta} = \{f_{\theta}, h^{ssl}_{\theta}, h^{cls}_{\theta}, h^{rot}_{\theta}\}$ is sequentially optimized on one task at a time up to the current one $t \in {1, ..., T_{c}}$. The objective function is therefore as follows:
\begin{equation}
    \mathcal{L}_{T_{c}} = \sum_{t=1}^{T_{c}} \displaystyle \mathop{\mathbb{E}}_{(x_{t}, y_{t}) \sim \mathcal{D}_{t}} \left[ l_{ce} (\sigma (h^{cls}_{\theta}(f_{\theta}(x_{t}))), y_{t}) \right],
\label{eqn_ce}
\end{equation}
where $\sigma$ is a softmax function and $l_{ce}$ is a classification loss, generally a cross-entropy loss. Continual learning is especially challenging since the data from the previous tasks are unavailable i.e at any point during training, the model $g_{\theta}$ has access to the current data distribution $\mathcal{D}_{t}$ alone. As the objective function in \cref{eqn_ce} is solely optimized for the current task, it leads to overfitting on the current task and catastrophic forgetting of older tasks. Replay-based methods sought to address this problem by storing a subset of training data from previous tasks and replaying them alongside current task samples. For replay-based methods, \cref{eqn_ce} can thus be rewritten as:
\begin{equation}
    \mathcal{L}_{cls} = \mathcal{L}_{T_{c}} + \displaystyle \mathop{\mathbb{E}}_{(x, y) \sim D_{r}} \left[ l_{ce} (\sigma (h^{cls}_{\theta}(f_{\theta}(x))), y) \right],
\label{eqn_er}
\end{equation}
where $D_{r}$ represents the distribution of samples stored in the buffer. Although cross-entropy loss is widely used for classification tasks in continual learning, it suffers from several shortcomings such as lack of robustness to noisy labels \citep{sukhbaatar2015training} and the possibility of poor margins \citep{poor_margin}, affecting the ability to generalize across tasks. As evidenced in \citep{gallardo2021selfsupervised}, self-supervised learning offers an alternative by learning task-agnostic, robust, and generalizable representations. We hypothesize that a two-stage training consisting of task-agnostic learning followed by task-specific learning can help bridge the aforementioned shortcomings without the need for pre-training. 

\begin{algorithm}[t]
\caption{The Proposed Method}
\label{alg:method}
\begin{algorithmic}[1]
\Statex \textbf{input: }training budget $\mathcal{B}$ and ratio $0<\gamma<1$, data streams $\mathcal{D}_{t}$ and $\mathcal{D}_{r}$
\ForAll {tasks $t \in \{1, 2,..,T\}$} 
    \For{ $e = 0 : \lfloor \gamma \mathcal{B} \rfloor$} \Comment{\textbf{Task-agnostic Learning}}
        \For{minibatch ${(X_{m}, Y_{m})_{m=1}^M} \in \mathcal{D}_{t} \cup \mathcal{D}_{r}$}
            \State Draw augmentation functions $a', a'' \sim A$
            \State $X = \{a'(X_m), a''(X_m)\}$
            \State $Z = h_{\theta}^{ssl}(f_{\theta}(X))$
            \State \begin{varwidth}[t]{\linewidth}
                  $\mathcal{L}_{ssl} =  \frac{1}{2\mathcal{N}}  \sum_{k=1}^{\mathcal{N}} [$ $l_{ssl}(2k-1, 2k) + l_{ssl}(2k, 2k-1)]$
                  \end{varwidth}
            \State Update the networks $f_{\theta}$ and  $h_{\theta}^{ssl}$
        \EndFor
    \EndFor
    \For{$e = \lfloor \gamma \mathcal{B} \rfloor : \mathcal{B}$} \Comment{\textbf{Task-specific Learning}}
        \For{minibatch ${(X_{m}, Y_{m})_{m=1}^M} \in \mathcal{D}_{t} \cup \mathcal{D}_{r}$}
            \State Draw rotation  $a \sim \{0^{0}, 90^{0}, 180^{0}, 270^{0}\}$
            \State $F = f_{\theta}(a(X_{m}))$
            \State $Z^{cls}, Z^{rot} = h_{\theta}^{cls}(F), h_{\theta}^{rot}(F)$ 
            \State Compute $L_{mo}$ 
            \State Update the networks $f_{\theta}$, $h_{\theta}^{cls}$ and $h_{\theta}^{rot}$
            \State Update replay buffer $\mathcal{D}_{r}$
        \EndFor
    \EndFor
\EndFor
\State \Return{model $f_\theta$ with $h_{\theta}^{cls}$}
\end{algorithmic}
\end{algorithm}

\subsection{Task-agnostic learning}
We aim to learn task-agnostic representations that are robust and generalizable across multiple tasks. Solving pretext tasks created from known information can help in learning representations useful for downstream tasks. Inspired by the recent advancements in contrastive self-supervised representation learning, we cast task-agnostic learning as an instance-level discrimination task. For a set of $\mathcal{N}$ randomly sampled images, each image is passed through two sets of augmentations $a', a'' \sim \mathcal{A}$ producing $2\mathcal{N}$ images per minibatch. Therefore, each image within $2\mathcal{N}$ samples will have a unique positive pair and $2(\mathcal{N}-1)$ negative samples. Let $Z = h_{\theta}^{ssl}(f_{\theta}(.))$ be a projection matrix of $2N$ augmented samples and $sim(., .)$ denote cosine similarity. The self-supervised contrastive loss \citep{simclr} for a positive pair of examples $(i, j)$ is defined as:
\begin{equation}
\label{eq:contrastive_loss}
l(i,j) =  - log\frac { { e }^{ sim(z_i,z_j)/\tau_{c}  } }{ \sum _{ k=1 }^{ 2\mathcal{N} }{ \mathbbm{1}_{[k \neq i]} { e }^{ sim(z_i,z_k)/\tau_{c}  } } }. 
\end{equation}

Contrastive learning in \cref{eq:contrastive_loss} learns visual representations by contrasting semantically similar (positive) and dissimilar (negative) pairs of data samples such that similar pairs have the maximum agreement via a contrastive loss in the latent space through Noise Contrastive Estimation (NCE) \citep{nce}. Given a limited training time for each task, it is pertinent to learn task-agnostic features that are in line with the class boundaries to avoid interference in the downstream tasks. Following \citet{supcontrast}, we adapt \cref{eq:contrastive_loss} to leverage label information. Within each minibatch, normalized embeddings belonging to the same class are pulled together while those belonging to other classes are pushed away in the latent space.  
\begin{equation}
\label{eq:supcon}
l_{ssl}(i,j) =  \frac{-1}{|\mathcal{P}(i)|}  \sum_{p \in \mathcal{P}(i)}  log\frac { { e }^{ sim(z_i,z_p)/\tau_{c}  } }{ \sum _{ k=1 }^{ 2\mathcal{N} }{ \mathbbm{1}_{[k \neq i]} { e }^{ sim(z_i,z_k)/\tau_{c} } } }, 
\end{equation}
where $\mathcal{P}(i)$ is a set of indices of samples belonging to the same class as the positive pair and $|\mathcal{P}(i)|$ is its cardinality. While \cref{eq:supcon} is a simple extension to contrastive loss, it eliminates the need for hard negative mining.

\subsection{Task-specific learning}
To align the task-agnostic representations to the current task, we train the classification head $h_{\theta}^{cls}$ with cross-entropy objective defined in \cref{eqn_er}. However, the interplay between task-agnostic and task-specific learning objectives can lead to sharp drift in the feature space and erode the generic representations learned during task-agnostic learning.

Multi-objective learning offers a viable solution to address this trade-off. Multi-objective learning can be thought of as a form of inductive transfer and is known to improve generalization \citep{Caruana}. It is also data efficient as multiple objectives are learned simultaneously using shared representations. However, simultaneous learning of multiple objectives poses new design and optimization challenges, and choosing which objectives that should be learned together is in itself a non-trivial problem \citep{crawshaw2020multitask}. We employ rotation prediction as an auxiliary loss to preserve the task-agnostic features. During task-specific stage of each task, input samples $x \in \mathcal{D}_{t} \cup \mathcal{D}_{r}$ are rotated by a fixed angle in addition to other transformations. The learning objective is to match task-specific ground truths $y \in \mathcal{D}_{t} \cup \mathcal{D}_{r}$ as well as auxiliary ground truths $y^a \in \{0^0, 90^0, 180^0, 270^0\}$, i.e. 
\begin{equation}
    \mathcal{L}_{mo} =  \alpha \mathcal{L}_{cls} + \beta \displaystyle \mathop{\mathbb{E}}_{x \sim \mathcal{D}_{t} \cup \mathcal{D}_{r}}  \left[ l_{ce} (\sigma (h^{rot}_{\theta}(f_{\theta}(x))), y^a) \right]
\label{eqn_dfdgf}
\end{equation}
where $\alpha$ and $\beta$ are hyperparameters for adjusting the magnitudes of two losses. \Cref{alg:method} summarizes the proposed method in detail.

\section{Experimental Results}

\begin{table*}
\centering
\caption{Comparison of CL models across various continual learning scenarios. We provide the average top-1 accuracy ($\%$) across all tasks after continual learning training.}
\begin{tabular}{ll|cccc|c|c}
\toprule
\multirow{2}{*}{\begin{tabular}[c]{@{}c@{}}Bufer\\ size\end{tabular}} & \multirow{2}{*}{Method} & \multicolumn{4}{c|}{Class-IL} & Domain-IL & General-IL \\
\cmidrule{3-8}
 & & CIFAR-10 & CIFAR-100 & TinyImageNet & STL-10 & R-MNIST & MNIST-360 \\ \midrule
 
\multirow{2}{*}{-}  & Joint  &  91.55 \scriptsize{$\pm$0.51} & 70.56 \scriptsize{$\pm$0.28} & 59.77 \scriptsize{$\pm$0.47} & 78.51 \scriptsize{$\pm$2.98} & 96.52 \scriptsize{$\pm$0.12} & 82.05 \scriptsize{$\pm$0.62}\\
& SGD & 19.55 \scriptsize{$\pm$0.06} & 17.49 \scriptsize{$\pm$0.28} & 8.00 \scriptsize{$\pm$0.14} & 17.35 \scriptsize{$\pm$1.52} & 70.76 \scriptsize{$\pm$5.61} & 21.09 \scriptsize{$\pm$0.21}  \\ \midrule

\multirow{2}{*}{-}  & oEWC  &  - & - & - & - & 76.06 \scriptsize{$\pm$4.51} & -\\
& oEWC + TARC & - & - & - & - & \textbf{89.57} \scriptsize{$\pm$1.18} & - \\ \midrule
\multirow{2}{*}{-}  & SI  &  - & - & - & - & 81.39 \scriptsize{$\pm$4.22} & -\\
& SI + TARC & - & - & - & - & \textbf{87.91} \scriptsize{$\pm$1.19} & -\\ \midrule
 
\multirow{2}{*}{200}  
  & ER  &  47.71 \scriptsize{$\pm$2.69} & 21.28 \scriptsize{$\pm$0.84} & 8.57 \scriptsize{$\pm$0.16} & 44.33 \scriptsize{$\pm$1.12} & 85.18 \scriptsize{$\pm$0.83} & 52.36 \scriptsize{$\pm$1.92}\\
& ER + TARC & \textbf{53.23} \scriptsize{$\pm$1.00} & \textbf{23.48} \scriptsize{$\pm$0.10} & \textbf{9.57} \scriptsize{$\pm$0.12} & \textbf{60.61} \scriptsize{$\pm$0.96} & \textbf{90.10} \scriptsize{$\pm$0.10} &  \textbf{69.54} \scriptsize{$\pm$1.94}\\ \midrule

\multirow{2}{*}{500}  
 & ER  & 61.97 \scriptsize{$\pm$1.77} & 27.85 \scriptsize{$\pm$0.39} & 10.33 \scriptsize{$\pm$0.24} & 58.07 \scriptsize{$\pm$1.06} & 88.03 \scriptsize{$\pm$0.94} & 66.27 \scriptsize{$\pm$2.82} \\
& ER + TARC & \textbf{67.41} \scriptsize{$\pm$0.94} & \textbf{31.50} \scriptsize{$\pm$0.40} & \textbf{13.77} \scriptsize{$\pm$0.17} & \textbf{67.96} \scriptsize{$\pm$1.84} & \textbf{90.00} \scriptsize{$\pm$0.27} & \textbf{72.25} \scriptsize{$\pm$3.25}\\ \midrule

\multirow{2}{*}{5120}  
& ER &\textbf{83.99} \scriptsize{$\pm$0.52} & \textbf{53.64} \scriptsize{$\pm$0.55} & 27.48 \scriptsize{$\pm$0.58} & 74.83 \scriptsize{$\pm$0.94} & \textbf{93.42} \scriptsize{$\pm$0.77} & 69.55 \scriptsize{$\pm$1.57}  \\
& ER + TARC & 82.21 \scriptsize{$\pm$0.38} & 52.30 \scriptsize{$\pm$0.20} & \textbf{32.04} \scriptsize{$\pm$0.37} & \textbf{75.28} \scriptsize{$\pm$1.06} & 92.29 \scriptsize{$\pm$0.28} & \textbf{73.01} \scriptsize{$\pm$1.74}\\ 
\bottomrule
\end{tabular}
\label{tab:main}
\end{table*}

\subsection{Empirical results}
\Cref{tab:main} provides a comparison of replay-based baseline ER \citep{Ratcliff1990, Robins1995CatastrophicFR} versus our method across various CL scenarios. We also provide a lower bound, \textit{SGD}, without resorting to any measures to address catastrophic forgetting and an upper bound, \textit{Joint}, where all tasks are trained together. 
We can make several observations from \cref{tab:main}: (i) Our method shows a strong performance in majority of the benchmarks, especially when the buffer size is small. We argue that our method is able to retain information more efficiently than ER by learning task-agnostic representations. (ii) In case of a larger buffer size, our method is behind ER by a small margin on smaller datasets. This is also true for other leading methods such as \citet{buzzega2020dark} where the gap in performance diminishes for large buffer sizes. 
(iii) Due to high class-to-buffer ratio (200/200, 200/500), TinyImageNet dataset is one of the hardest benchmarks under Class-IL setting. Our method outperforms the baseline across all buffer sizes. (iv) The relative improvement compared to the baseline improves with the increase in image size. Task-agnostic learning is able to discriminate better for STL-10 and TinyImageNet when the images are comparatively larger in size.

The benefits our training paradigm is not limited to replay-based methods alone. To prove the generalizability of our method, we adapt two regularization-based approaches (oEWC \citep{schwarz2018progress} and SI \citep{si}) and provide their results in Domain-IL scenario. We employ the same training schedule as in \cref{alg:method} except that the number of epochs is limited to 2 and task-specific learning also includes a regularization term both in oEWC and SI. We also do not use any replay for these methods. R-MNIST under Domain-IL consists of 20 subsequent tasks and is more challenging dataset as it tests forward facilitation across longer task sequence. On R-MNIST, Our method outperforms the baseline by a large margin indicating that our framework works across different CL approaches. 

\Cref{tab:main} also presents the comparison of ER and TARC in more challenging General-IL scenario. It is worth noting that General-IL setting presents both sharp (changes in class) and smooth distribution (rotation) shifts, and involves recurring classes in subsequent sequences which makes the transfer of knowledge from previous occurrences important. Our method is able to leverage positive transfer when revisiting the previous task and outperforms the ER across different buffer sizes. Similar to prior results, the improvement is highest among low buffer regimes indicating the superiority of our method.

\subsection{Forward and backward transfer}
Following \citet{gem}, we compute forward transfer as the difference between the accuracy just before starting training on a given task and the one of the random-initialized network. While one can argue that learning to classify unseen classes is desirable, Class-IL shows distinct classes in distinct tasks, which makes transfer impossible. On the contrary, forward transfer can be relevant for Domain-IL scenarios, provided that the input transformation is not disruptive. As is the case with the R-MNIST, it requires the CL model to classify all digits for 20 subsequent tasks, with images rotated by a random angle in the interval $[0, \pi)$. Therefore, the positive forward transfer is not only plausible but also a highly desirable property in this setting. As far as the backward transfer is concerned, we compute the difference between the current accuracy and its best value for each task. Table \ref{tab:fwt} presents the forward and backward transfer results on R-MNIST, averaged across all task. As can be seen, TARC exhibits strong forward transfer and improved backward transfer when compared to the respective baselines.

\begin{table}[t]
 \center
\caption{Forward transfer and backward transfer on R-MNIST dataset.}
\label{tab:fwt}
    \begin{tabular}{l|l|cc||l|cc}
    \toprule
    \begin{tabular}[c]{@{}l@{}}Buffer \\ size\end{tabular}  & Method & \begin{tabular}[c]{@{}l@{}}Forward \\ Transfer\end{tabular} & \begin{tabular}[c]{@{}l@{}}Backward \\ Transfer\end{tabular} & Method & \begin{tabular}[c]{@{}l@{}}Forward \\ Transfer\end{tabular} & \begin{tabular}[c]{@{}l@{}}Backward \\ Transfer\end{tabular} \\ 
    \midrule
    \multirow{2}{*}{200} & ER & 62.33\scriptsize{$\pm$4.47} & -12.30 \scriptsize{$\pm$0.89} & oEWC & 52.41\scriptsize{$\pm$6.24}  & -21.55 \scriptsize{$\pm$4.68} \\
    & ER + TARC & \textbf{75.65}\scriptsize{$\pm$2.06} & \textbf{-1.37}\scriptsize{$\pm$0.09} & oEWC + TARC & \textbf{74.05}\scriptsize{$\pm$2.02} & \textbf{-1.56} \scriptsize{$\pm$1.22}\\
    \midrule
    \multirow{2}{*}{500} & ER & 65.92\scriptsize{$\pm$3.02} & -9.26\scriptsize{$\pm$1.04} & SI & 	52.79 \scriptsize{$\pm$6.76}  & 	-17.82 \scriptsize{$\pm$3.75} \\
    & ER + TARC & \textbf{76.01}\scriptsize{$\pm$1.52} & \textbf{-1.51}\scriptsize{$\pm$0.20} & SI + TARC &  	\textbf{71.81} \scriptsize{$\pm$0.99} & \textbf{-1.40} \scriptsize{$\pm$1.09} \\
    \midrule
    \multirow{2}{*}{5120} & ER	& 73.07\scriptsize{$\pm$1.37} & -3.61\scriptsize{$\pm$0.73} \\
    & ER + TARC & \textbf{78.79}\scriptsize{$\pm$1.50} & \textbf{-0.11}\scriptsize{$\pm$0.14} \\
    \bottomrule
    \end{tabular}
\end{table}


\begin{figure}[t]
\centering
  \includegraphics[width=\linewidth]{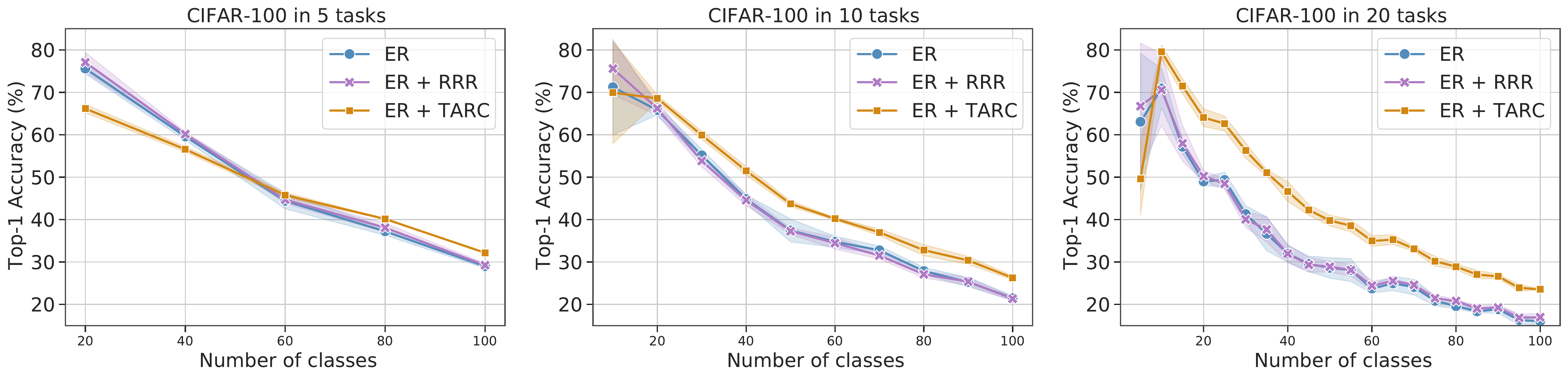}
  \caption{Comparison of our method against RRR across different number of tasks in CIFAR-100.}\label{fig:rrr}
\end{figure}

\subsection{Comparison with RRR}
Similar to our method, RRR proposed a training paradigm that is extensible to existing CL approaches. \Cref{fig:rrr} presents comparison of RRR against our method on CIFAR-100 under different number of task sequences. Following RRR, we tweak our training schedule and update the replay buffer once after each task. Gradient Class Activation Mapping \citep{gradcam} is used for generating explanations for the buffered images. As the length of task sequence increases, our method clearly outperforms both RRR and ER. Our method is able to consolidate the generalizable features better and mitigate the effect of catastrophic forgetting. Unlike pre-training, the effect of task-agnostic learning does not diminish with longer task sequences. Since both TARC and RRR are generic frameworks, a combination of them  might improve the our proposed method even further.

\section{Model Analyses}
\label{analyses}
We attempt to gain insights into the working mechanism of our method and elaborate on the additional advantages it brings without any explicit constraints.

\begin{figure}[t]
\centering
\includegraphics[trim=0 7 0 7 cm, width=0.9\linewidth]{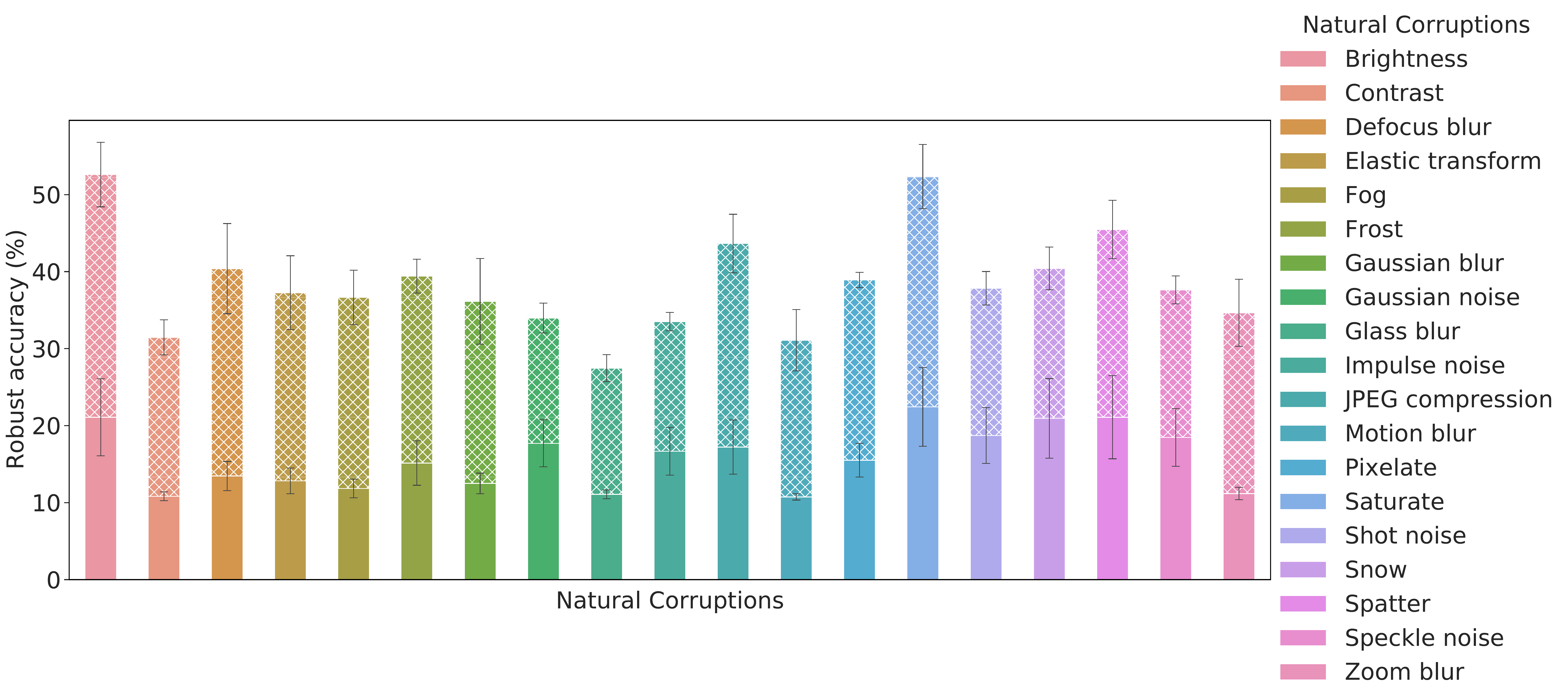}
\caption{Robustness to natural corruptions. The unshaded part of each bar represents the robust accuracy of ER against 19 natural input corruptions. The shaded part of each bar represents the improvement our method {(ER + TARC)} achieves over the ER baseline.}
\label{fig:corruptions}
\end{figure}

\begin{figure}[tb]
  \centering
  \begin{tabular}{cc}
  \includegraphics[width=0.48\linewidth, height=0.25\textwidth, keepaspectratio]{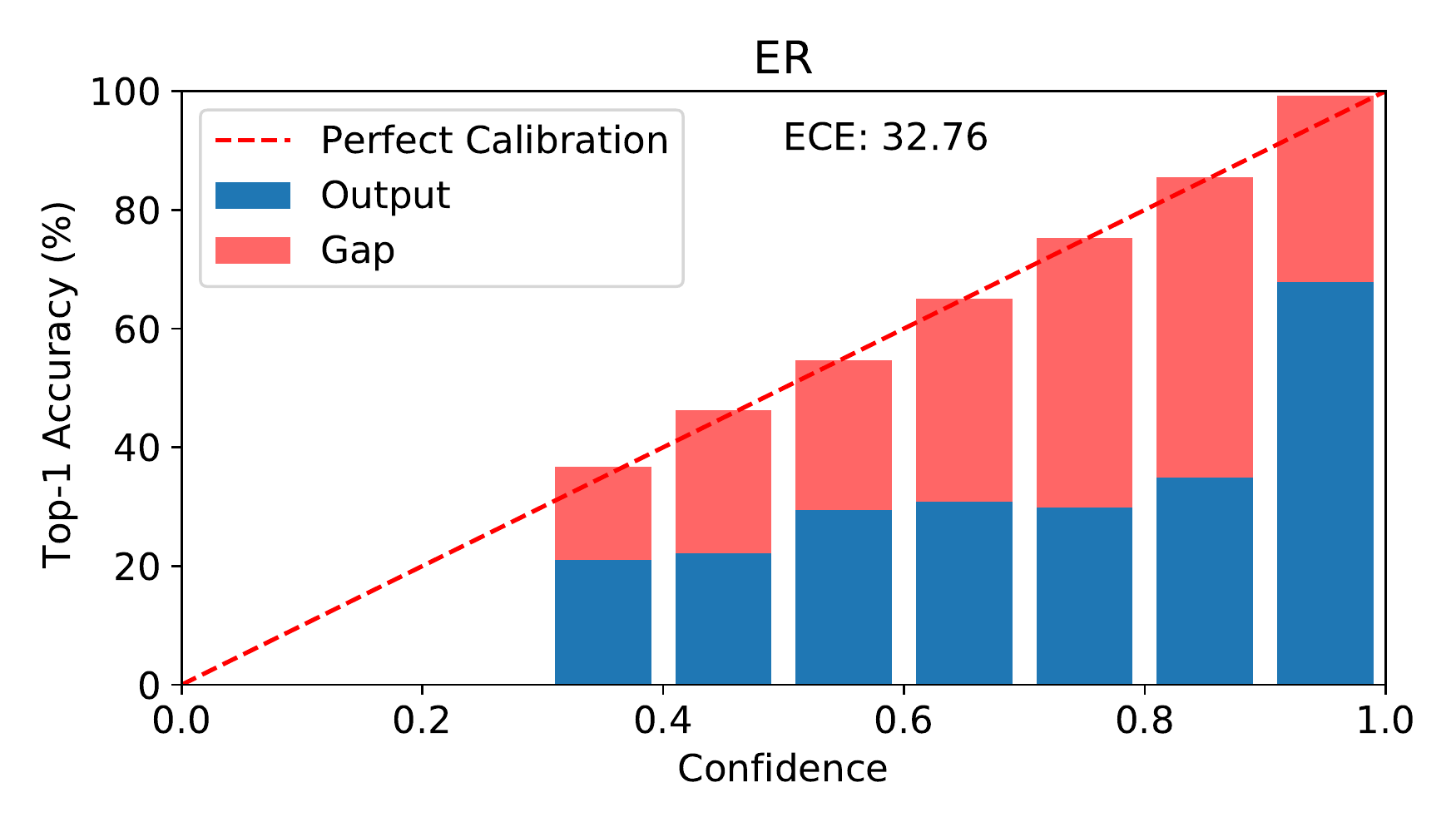} &
  \includegraphics[width=0.48\linewidth,  height=0.25\textwidth, keepaspectratio]{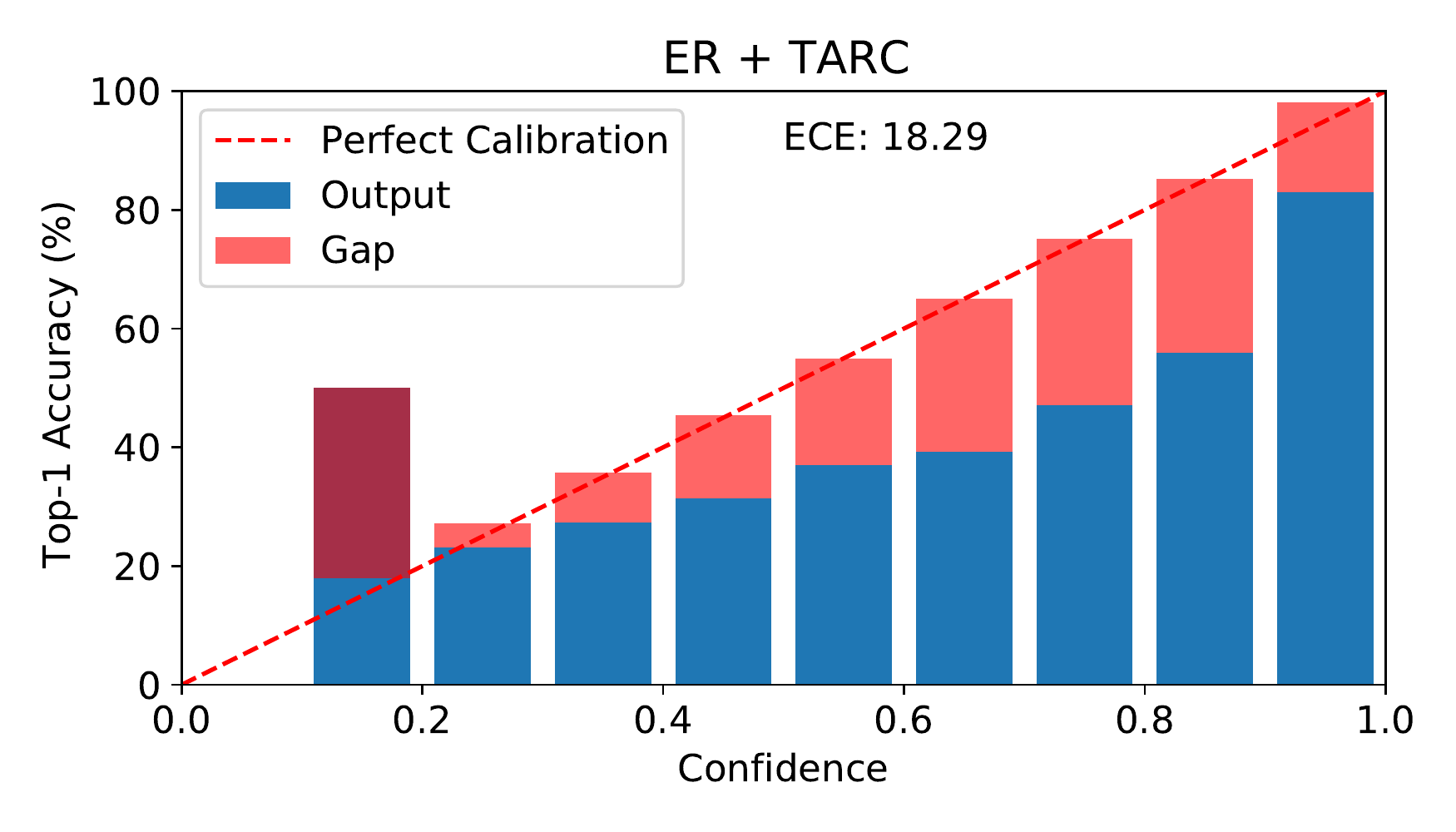}
  \end{tabular}
  \caption{Confidence estimates and corresponding Expected Calibration Error (ECE) of CIFAR-10 trained CL models. Lower ECE is better. Our method is well calibrated with confidence estimates closer to the perfect calibration compared to ER.}
\label{fig:calibration}
\end{figure}

\textbf{Robustness to natural corruptions}: Autonomous CL agents operating in the real world are exposed to ever changing environments, often influenced by illumination and weather changes. Therefore, it is pertinent for CL agents to be robust to data distributions with natural corruptions. In this section, we evaluate our method against common image corruptions using CIFAR-10-C \citep{hendrycks2019}. Models are trained using clean CIFAR-10 with a buffer size of 500 and tested on CIFAR-10-C. Mean Corruption Error (mCE) and Relative Mean Corruption Error (Relative-mCE) are commonly used metrics to evaluate the performance under natural corruptions \citep{hendrycks2019}. \Cref{fig:corruptions} shows robustness to 19 different corruptions averaged over five severity levels. The shaded region in each bar in \cref{fig:corruptions} indicates an improvement over the baseline (ER). Our method has a lower mCE across all different corruptions, achieving $61.53\%$ while ER achieves  $84.24\%$. When comparing against their respective natural accuracies, Relative-mCE is a better measure to compare models with different top-1 accuracy. Our method achieves $28.94\%$ Relative-mCE while ER settles for $46.21\%$. Evidently, task-agnostic learning when coupled with task-specific learning brings discernible benefits in terms of robustness to natural corruptions. 

\begin{figure}[t]
\centering
\includegraphics[width=\linewidth, keepaspectratio]{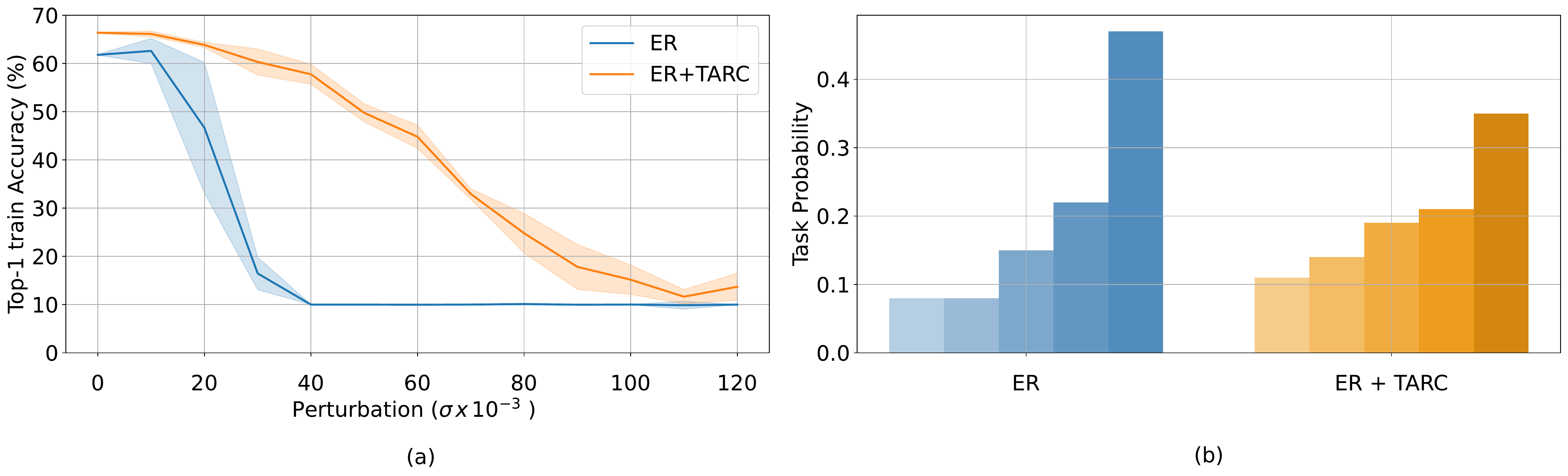}
\caption{(a) Robustness of CL models to varying degrees of  Gaussian noise added to the model weights. Our method is considerably robust to Gaussian perturbations and loses performance gradually suggesting convergence to flatter minima. (b) Average task probabilities of CL models trained on CIFAR-10 with 500 buffer size. Within each bar group, right most bar indicates the most recent task. Our method reduces the bias towards most recent task.}
\label{fig:gauss_noise}
\end{figure}


\begin{figure}[t]
\centering
\includegraphics[width=\linewidth]{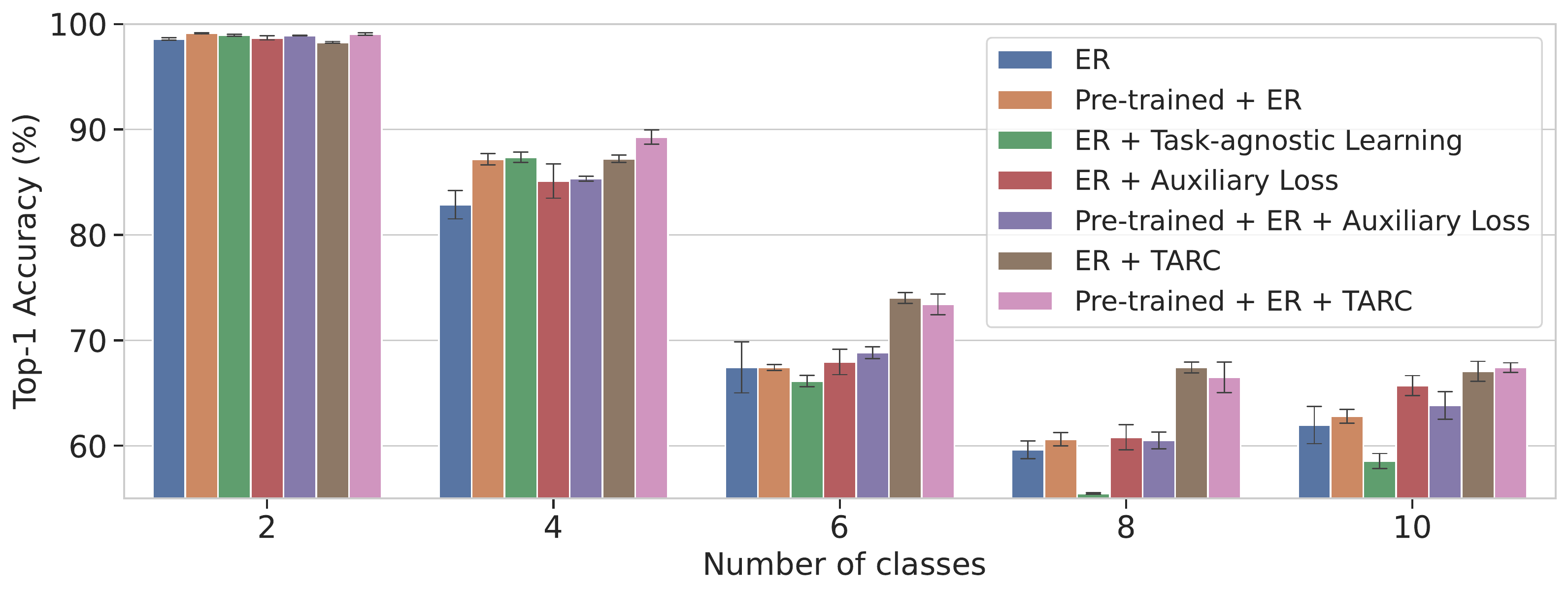}
\caption{Effect of different components added to ER. TARC has the highest performance gain compared to individual components over ER. As number of tasks increases, pre-training has little to no effect on different CL methods.}
\label{fig:abllation_extended}
\end{figure}

\textbf{Model calibration}: In safety-critical applications, it is pertinent for a model to possess an adequate sense of uncertainty about its predictions. 
 A model is said to be miscalibrated when it tends to be overconfident or under-confident about its predictions compared to ground truth accuracy \citep{pmlr-v70-guo17a}. In this section, we evaluate how well the models are calibrated. Expected Calibration Error (ECE) \citep{naeini2015obtaining} is the most common metric to determine miscalibration in classification. ECE computes a weighted average over the difference between the absolute accuracy and average confidence. A lower ECE value indicates a better calibrated model. We report ECE along with a reliability diagram using a calibration framework \citep{calibration}. 
  \Cref{fig:calibration} shows the reliability diagram of our method versus the baseline. Result shows that ER predictions are more overconfident compared to our method resulting in ECE values $32.76$ and $18.29$ respectively. In addition to improvement in natural accuracy, the right combination of task-agnostic and task-specific learning can effectively improve calibration, thereby improving reliability in safety critical environments.

\textbf{Convergence to flatter minima}: A CL model that converges to flatter minima in a loss landscape has more flexibility to adapt to a new task without drifting too far from the optimal parameters for the previous tasks. Furthermore, the solutions that reside in a flatter minima are more robust as the predictions do not change drastically with small perturbations \citep{buzzega2020dark}. 
Following the analysis in \citep{Zhang_2018_CVPR, buzzega2020dark}, we add independent Gaussian noise to all parameters of the model trained on CIFAR-10 with 500 buffer size. \Cref{fig:gauss_noise}-(a) shows the change in accuracy over  different perturbation levels for the training set. Compared to the ER baseline, our method is significantly less sensitive to perturbations and the performance drops smoothly. We argue that task-agnostic learning followed by task-specific learning guides the solution to a wider valley which could better explain the ability of our model to consolidate generalizable features.

\textbf{Bias towards recent tasks}: Due to the sequential nature of continual learning, predictions are biased towards the recent task as the number of samples for the current task are far more than the buffered samples. Explicit techniques such as cosine normalization \citep{Hou_2019_CVPR}, weight aligning \citep{zhao2019maintaining} have been employed in the past to reduce the bias towards recent tasks. 
Following the analysis in \citep{buzzega2020rethinking}, the normalized probability of each task of a CIFAR-10 trained model is computed by averaging probabilities of all samples belonging to the associated classes in Class-IL setting. \Cref{fig:gauss_noise}-(b) shows the normalized probability of each task being predicted at the end of training. Our method reduces the bias towards the most recent task and task probabilities are more evenly distributed compared to ER. Intertwining of task-agnostic learning and task-specific learning implicitly reduces this bias without any additional constraints as the model spends more time on learning generalizable features than aligning itself with the current task.


\begin{figure}[t]
    \centering
    \begin{minipage}{0.33\textwidth}
        \centering
        \includegraphics[width=1\linewidth]{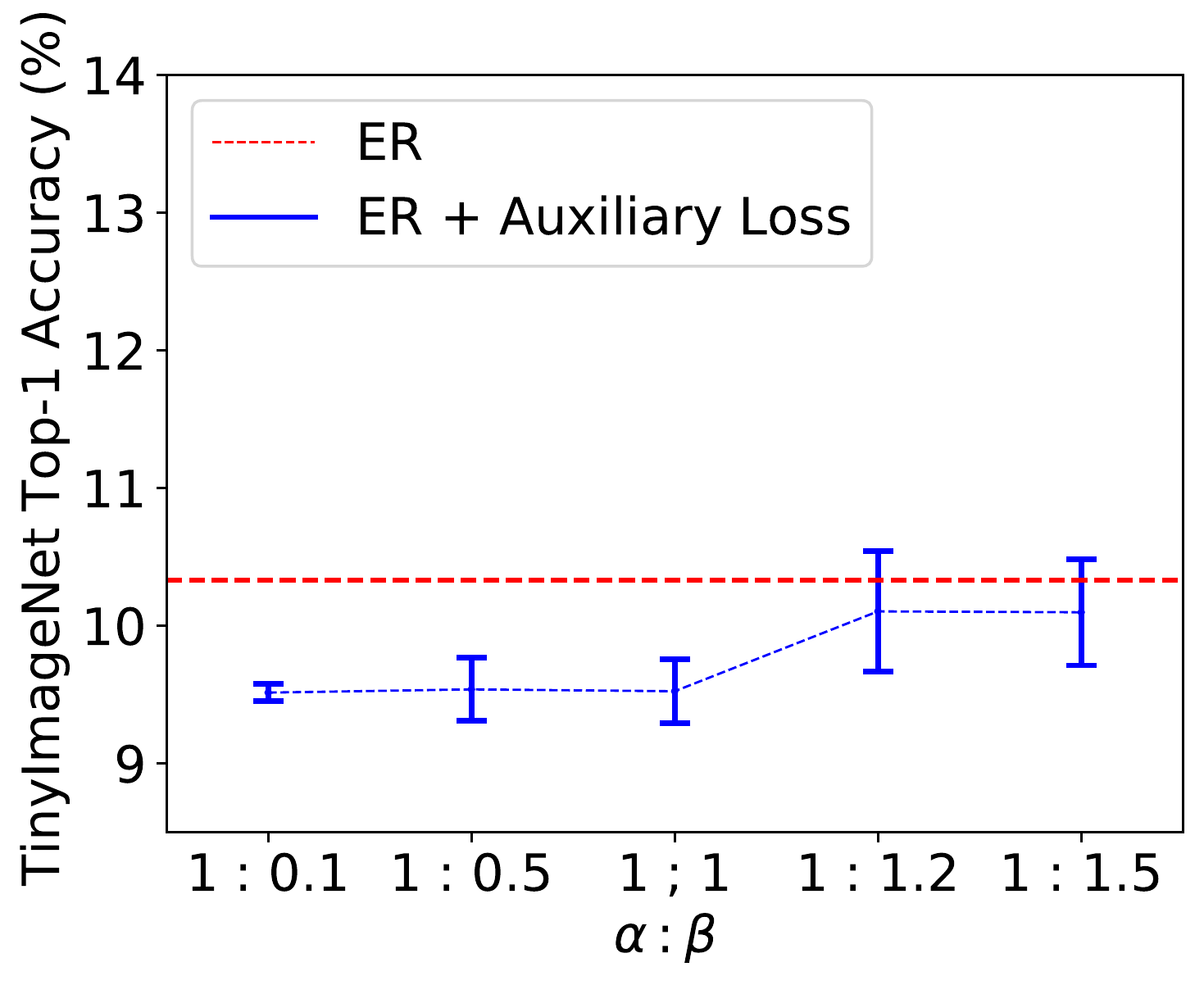} \\
        $(a)$
    \end{minipage}%
    \begin{minipage}{0.33\textwidth}
        \centering
        \includegraphics[width=1\linewidth]{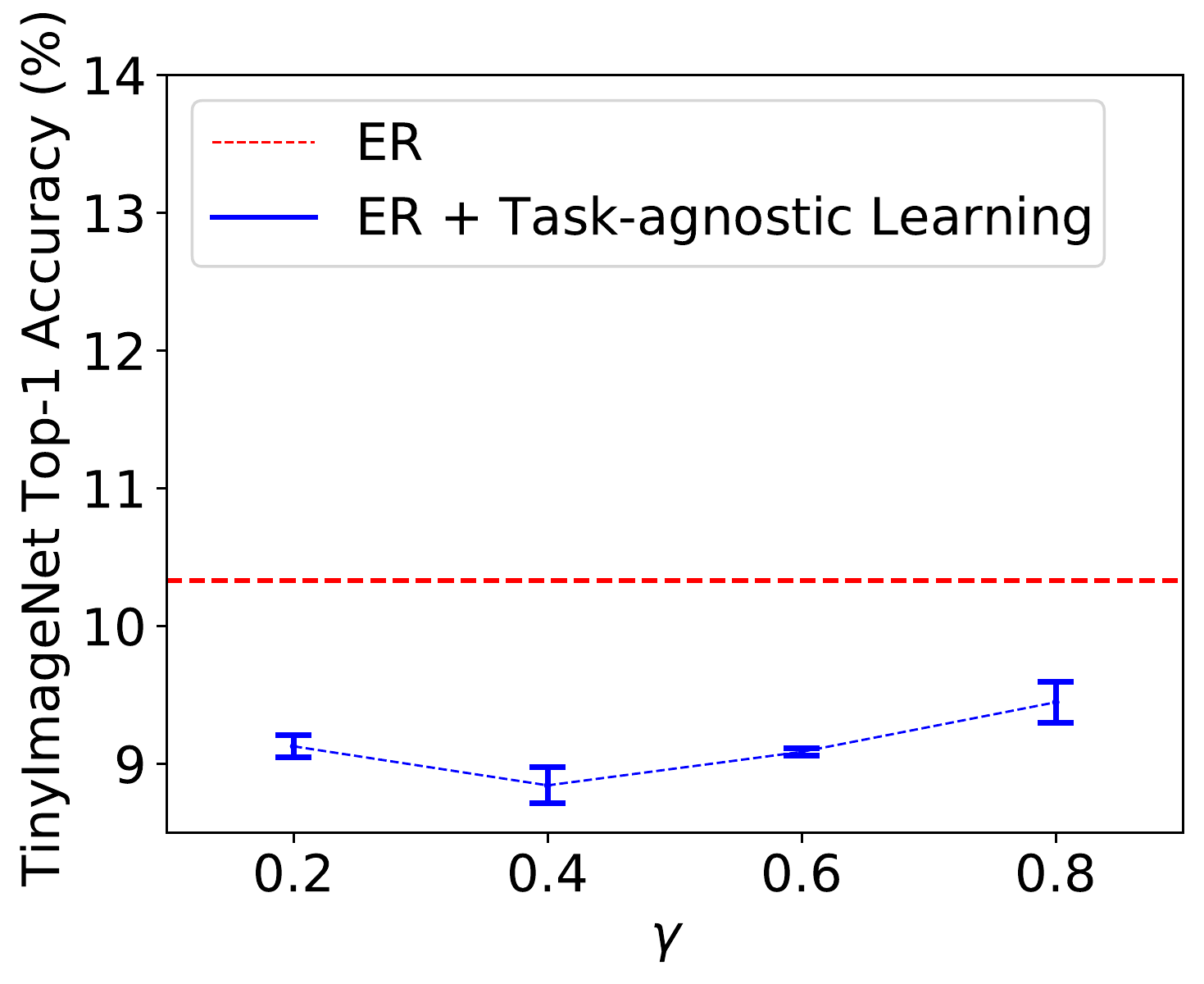} \\
        $(b)$
    \end{minipage}
        \begin{minipage}{0.33\textwidth}
        \centering
        \includegraphics[width=1\linewidth]{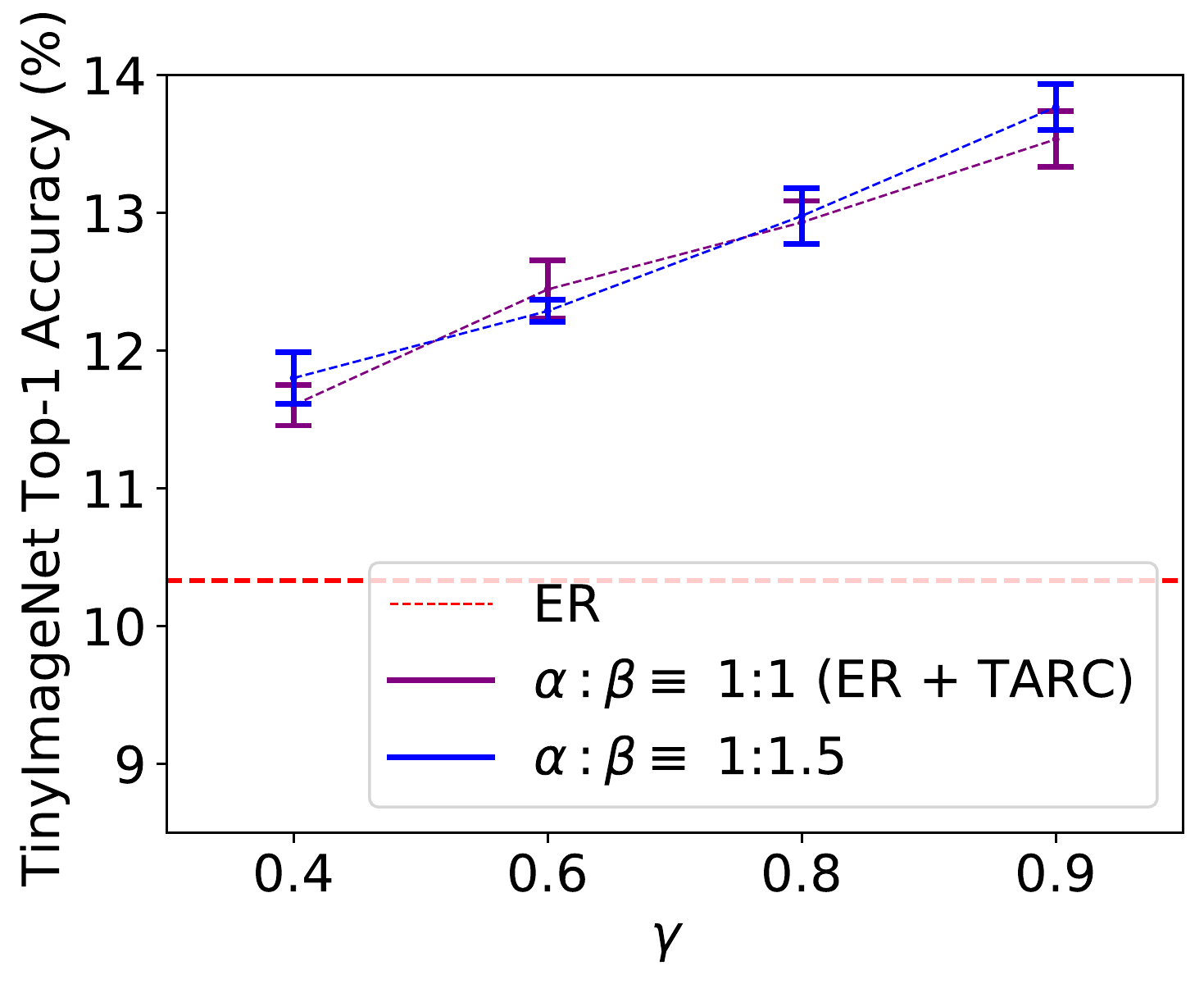} \\
        $(c)$
    \end{minipage}
 \caption{Hyperparmeter tuning for optimal CL performance. Graphs $(a)$ and $(b)$ consider our two main components of TARC separately with ER. As can be seen, the results are not satisfactorily above the ER baseline. The right most graph $(c)$ presents our method TARC with different values for the hyperparamters. The results reaffirm our earlier hypothesis that both task-agnostic learning and auxiliary loss are essential components of our method.}
\label{fig:ablation1}
\end{figure}

\section{Ablation Study} \label{ablation_study_section}
We attempt to shed light on the working of each component in our method. One might wonder whether task-agnostic learning or multi-objective learning alone can reap the aforementioned benefits for CL. We hypothesize that both task-agnostic learning and auxiliary loss are critical for our method since the former helps in learning generalizable features while the latter reduces the risk of these features being overwritten from task-specific learning. \Cref{fig:abllation_extended} presents the evaluation of different variants of ER. We also extend the ablation study with more combinations of pre-training, auxiliary loss, and TARC. In line with our understanding, task-agnostic learning and auxiliary loss complement each other when coupled together. Our method is computationally less expensive and superior to pre-training on TinyImageNet. As number of tasks increases, pre-training has little to no effect on different CL methods. On the contrary, TARC provides discernible performance improvement over all other combinations with/without pre-training. We argue that this performance improvement over pre-training is essentially due to our method's ability to diminish the domain shift.


\section{Hyperparameter tuning}
We explore the effect of hyperparameters $\alpha$, $\beta$ and $\gamma$ on our method. \Cref{fig:ablation1} presents a study on the effect of these parameters on the the performance of our method on TinyImageNet. To reap the full benefits of our method, we note that it is crucial to have both task-agnostic learning and auxiliary loss paired with the baseline. As can be seen from the graph, when these components are used in isolation with ER, they do not perform satisfactorily above the ER baseline. Contrary to earlier methods that show nominal improvement over the ER baseline by using  one of these components, TARC shows significant performance improvement by effectively combining both these components.  Unless otherwise specified, the optimal parameters $1$, $1$ and $0.9$ are used for $\alpha$, $\beta$ and $\gamma$ respectively in all the experiments presented in this paper.


\section{Conclusion and Future work}
We proposed a novel two-stage CL training paradigm that intertwines task-agnostic and task-specific learning. Unlike self-supervised pre-training, our method does not suffer from domain shift, non-availability  of  pre-training  data  and  additional computational costs. Furthermore, the effect of task-agnostic learning in our method does not diminish with longer task sequences. Our method can be easily added to memory- or regularization-based approaches with consistent performance gain across more challenging CL settings. We further provided an extensive analyses to shed light on the superiority of our method in terms of robustness, model calibration and bias towards recent tasks. Extending our method to algorithms oblivious of task boundaries and/or knowledge distillation based approaches are some of the useful future research directions for this work.

\bibliography{egbib}
\bibliographystyle{collas2022_conference}

\appendix

\section{Experimental setup}
 Following \citep{hsu2019reevaluating, buzzega2020dark}, we evaluate on the following CL scenarios:
 
\textbf{Class Incremental Learning (Class-IL)}: In this setting, the model encounters a new set of classes in each task and must learn to distinguish all classes encountered thus far after each task. In practice, we split CIFAR-10 \citep{cifar10}, CIFAR-100 \citep{cifar10}, TinyImageNet \citep{tinyimagenet}, and STL-10 \citep{stl10} into partitions of 2, 20, 20, and 2 classes per task, respectively.  


\textbf{Domain Incremental Learning (Domain-IL)}: In this setting, the number of classes remain the same across subsequent tasks. However, a task-dependent transformation is applied changing the input distribution for each task. Specifically, R-MNIST \citep{gem} rotates the input images by a random angle in the interval $[0; \pi)$. R-MNIST requires the model to classify all 10 MNIST \citep{mnist} digits for 20 subsequent tasks.


\textbf{General Incremental Learning (General-IL)}: In this setting, MNIST-360 \citep{buzzega2020dark} models a stream of MNIST data with batches of two consecutive digits at a time. Each sample is rotated by an increasing angle and the sequence is repeated six times. General-IL exposes the CL model to both sharp class distribution shift and smooth rotational distribution shift.

\section{Implementation details}
To ensure a fair comparison between different methods in incremental learning, we build upon the \textit{mammoth} CL framework \citep{buzzega2020dark} in PyTorch. Given a training budget $\mathcal{B}$ for each task, we adapt the original training schedule to accommodate self-supervised learning and multi-objective learning during the CL. Rotation prediction is used as an auxiliary task alongside supervised learning during multi-objective learning. 

We employ ResNet-18 \citep{resnet18} for Class-IL and a 2-layer fully-connected network of 100 neurons each with ReLU activation for Domain-IL tasks. The underlying network consists of three different heads $h^{ssl}_{\theta}, h^{rot}_{\theta}$, and $h^{cls}_{\theta}$ each for supervised contrastive learning, rotation prediction and classification. $h^{ssl}_{\theta}$ consists of a linear layer followed by a projection head of two-layer MLP with a ReLU non-linearity and 1-dimensional batch norm. We use ADAM \citep{kingma2017adam} optimizer with a learning rate of $3e^{-4}$. During task-agnostic learning, input images are transformed using a stochastic augmentation module consisting of a random resized crop, random horizontal flip followed by random color distortions. Since images are smaller in size, we leave out gaussian blur. 
During multi-objective learning, input images are rotated by one of $\{0^0, 90^0, 180^0, 270^0\}$ degrees. A linear layer is used for each rotation prediction $h^{rot}_{\theta}$ and classification $h^{cls}_{\theta}$ with cross entropy as a learning objective. 

\section{Evaluation under noisy labels}
\begin{figure}[h]
\centering
\includegraphics[width=0.4\linewidth, keepaspectratio]{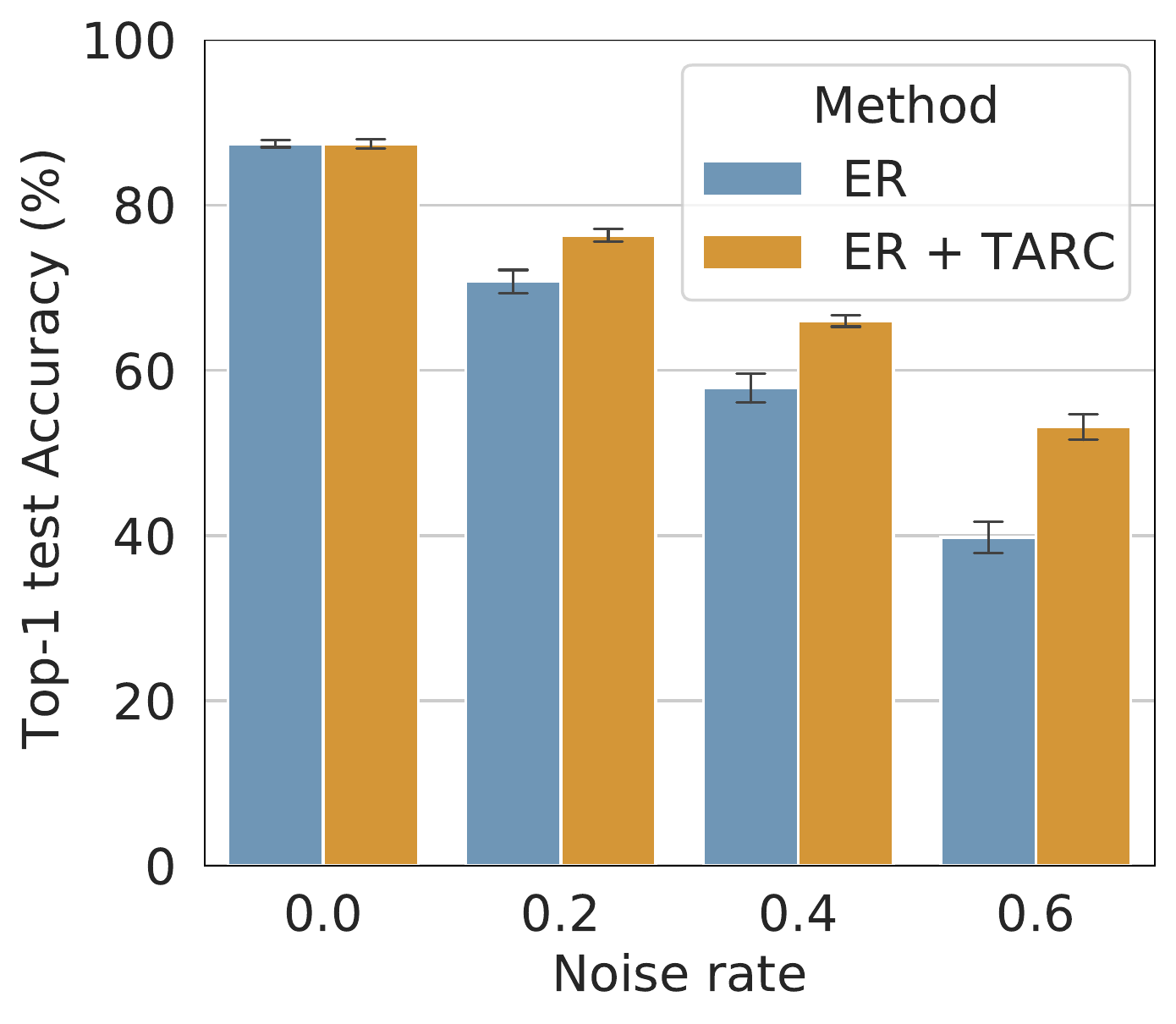}
\caption{Robustness of CL models trained on CIFAR-10 with 500 buffer size against symmetric noisy labels. }\label{fig:noisy_labels}
\end{figure}

Given the limited training budget for each task, quality of annotations play a crucial role in the success of CL methods. Robust training procedures need to be put in place to  offset the impact of noisy labels as deep neural networks are known to memorize noisy labels \citep{arpit2017closer}. We hypothesize that the representations learned using our method are more robust to noisy labels. To test our hypothesis, we simulate label corruption on CIFAR-10 by randomly sampling labels from a uniform distribution with a given probability (noise rate). A linear layer is trained on top of the backbone $f_{\theta}$ in presence of noisy labels and evaluated on the clean test set. \Cref{fig:noisy_labels} presents robust accuracy  for different noise rates. our method is less sensitive to noisy labels across different noise rates. By intertwining task-agnostic and task-specific learning, our method decouples representation learning from classifier. Since task-agnostic learning dominates the majority of the training budget $\mathcal{B}$, CL model has less chance to overfit to the noisy labels. 


\section{Analyses of our method on oEWC}
We extend some of the analyses in section 5 to oEWC. We use the same mechanism explained in the Algorithm 1 and extend oEWC to Domain-IL setting. As pointed out earlier, our method outperforms the oEWC baseline by almost $13\%$ in Domain-IL setting. 

\subsection{Model calibration}
\begin{figure}[!htb]
    \centering
    \begin{minipage}{.5\textwidth}
        \centering
        \includegraphics[width=1\linewidth, height=1\textheight, keepaspectratio]{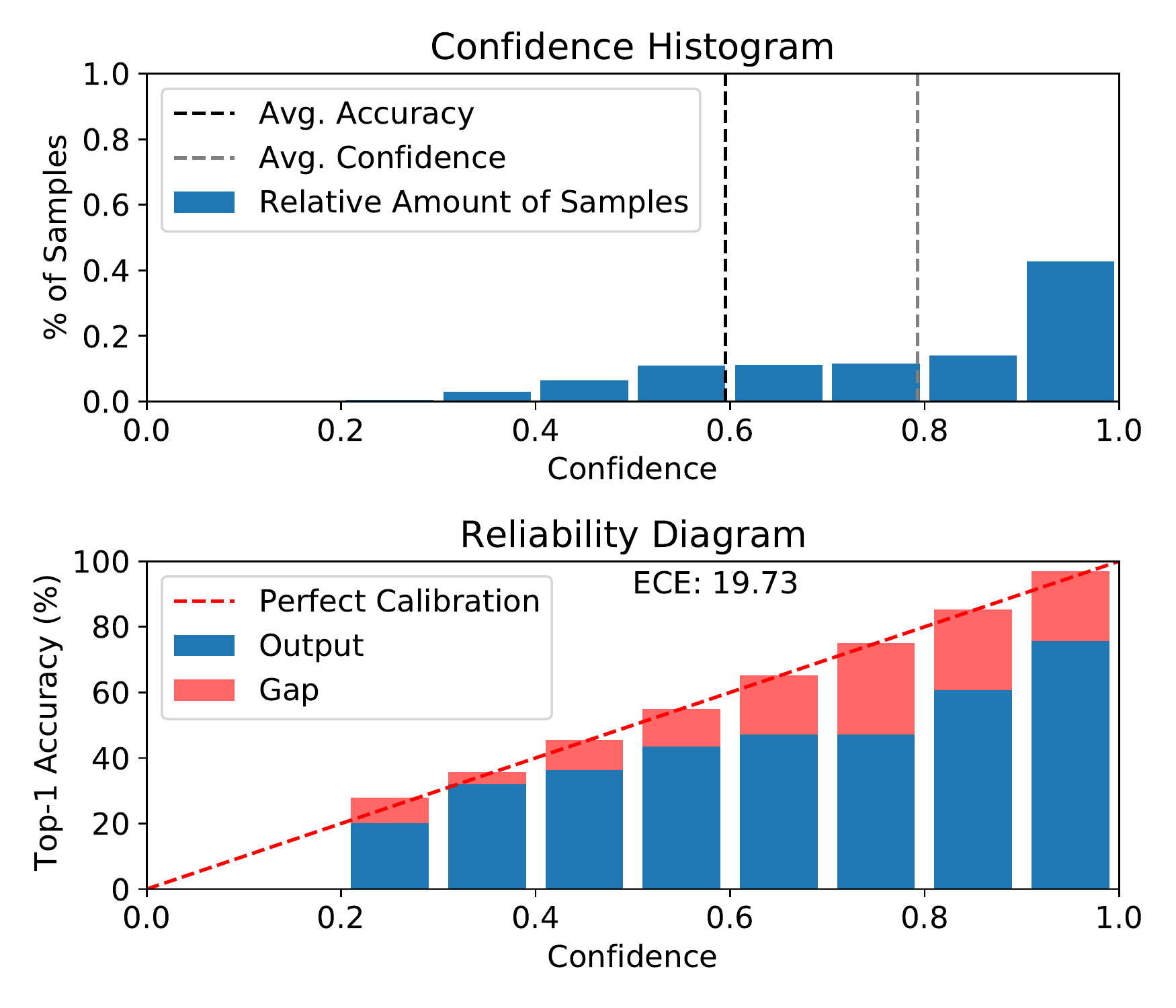} \\
        $(a)$ Model calibration - oEWC
    \end{minipage}%
    \begin{minipage}{0.5\textwidth}
        \centering
        \includegraphics[width=1\linewidth, height=1\textheight, keepaspectratio]{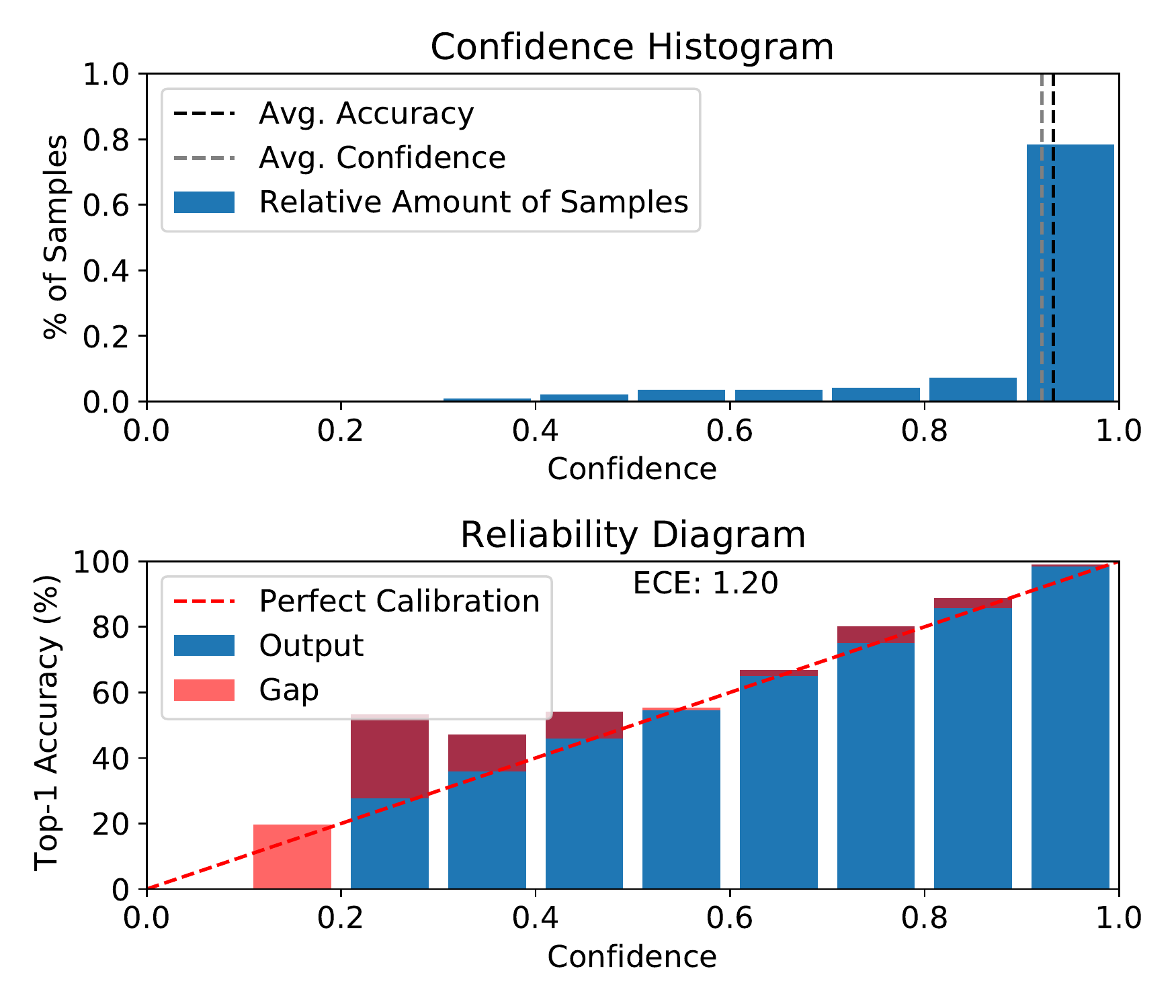} \\
        $(b)$ Model calibration - oEWC + TARC
    \end{minipage}
  \caption{Confidence estimates and corresponding ECE of R-MNIST trained CL models on clean MNIST. Lower ECE is better. Our method is well calibrated with confidence estimates closer to the perfect calibration compared to oEWC.}
  \label{fig:oewc_calibration}
\end{figure}

A model is said to be miscalibrated when it tends to be overconfident or under-confident about its predictions compared to ground truth accuracy. \Cref{fig:oewc_calibration} presents the reliability diagram of our method versus the baseline. Result shows that ER predictions are more overconfident compared to our method resulting in ECE values 19.73 and 1.20 respectively.  In addition to improvement in natural accuracy over $13\%$ in Domain-IL setting, TARC  effectively improves model calibration, thereby improving reliability in safety critical environments.


\subsection{Flatter minima analysis}
\begin{figure*}[hbt!]
\centering
\includegraphics[width=0.4\linewidth]{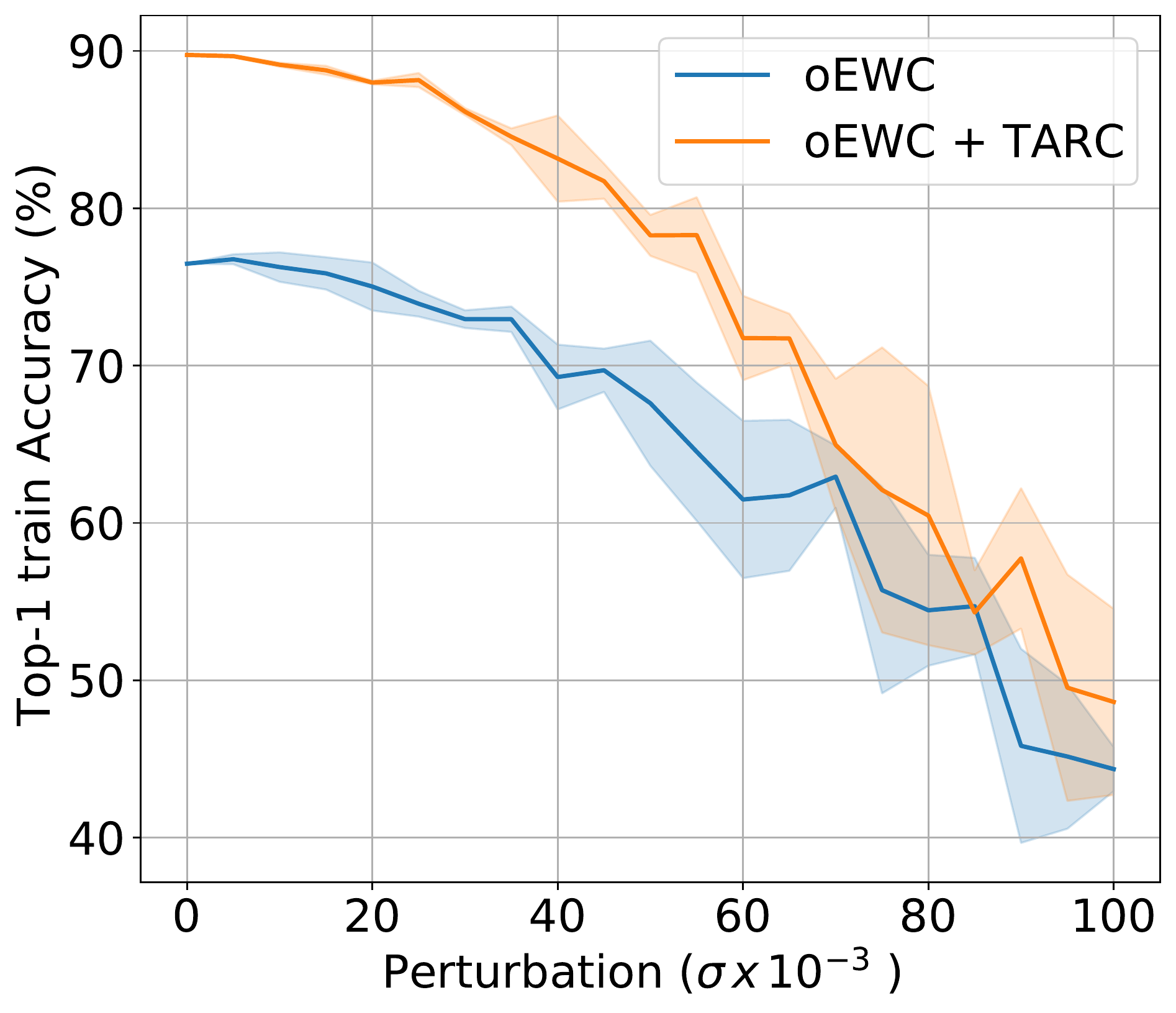}
\caption{Robustness  of  CL models  trained on R-MNIST to  varying  degrees  of Gaussian   noise   added   to   the model  weights. Our  method is considerably robust to Gaussian perturbations and loses performance  gradually  suggesting convergence to flatter minima. }
\label{fig:gauss_oewc}
\end{figure*}

We extend the analysis conducted in section 5 to Domain-IL scenario. We add independent noise to all parameters of models trained on R-MNIST. Figure \ref{fig:gauss_oewc} presents change in accuracy over different perturbation levels for MNIST training set. Compared to oEWC baseline, our method is significantly less sensitive to perturbations and loses performance gradually suggesting convergence to a flatter minima in the loss landscape. 

\section{Self-supervised pre-training for continual learning}
Table \ref{tab:pretraining} provides a comparison of the proposed training paradigm with pre-trained baselines across different datasets in Class-IL scenario. Pre-training is carried out on TinyImageNet using SimCLR \cite{simclr}, a self-supervised learning objective. Similar to our main results in Table \ref{tab:main}, our method shows a strong performance in majority of the benchmarks, especially when the buffer size
is small. We argue that the effect of pre-training diminishes with the onset of CL training. On the other hand, TARC is able to retain information more efficiently than pre-training baselines by learning task-agnostic representations during CL training. In case of larger buffer size, our method is behind pre-trained baselines by a small margin on smaller datasets. The relative improvement compared to the baseline improves with the increase in image size and longer task sequences. TARC is able to discriminate better for STL-10 and TinyImageNet when the images are comparatively larger in size. For buffer size 500, the relative improvement in CIFAR-10 is close to $8\%$ while it is approximately $35\%$ in TinyImageNet.

\begin{table*}
\centering
\caption{Comparison of TARC and SSL pre-training on CL models across various scenarios. We provide the average top-1 accuracy ($\%$) across all tasks after continual learning training.}
\begin{tabular}{ll|cccc}
\toprule
\multirow{2}{*}{\begin{tabular}[c]{@{}c@{}}Bufer\\ size\end{tabular}} & \multirow{2}{*}{Method} & \multicolumn{4}{c}{Class-IL} \\
\cmidrule{3-6}
 & & CIFAR-10 & CIFAR-100 & TinyImageNet & STL-10 \\ \midrule

\multirow{2}{*}{200}  & Pretrained + ER  &  49.27 \scriptsize{$\pm$0.87} & 22.57 \scriptsize{$\pm$0.39} & 8.67 \scriptsize{$\pm$0.10} & 56.71 \scriptsize{$\pm$0.62} \\

& ER + TARC & \textbf{53.23} \scriptsize{$\pm$1.00} & \textbf{23.48} \scriptsize{$\pm$0.10} & \textbf{9.57} \scriptsize{$\pm$0.12} & \textbf{60.61} \scriptsize{$\pm$0.96} \\ \midrule

\multirow{2}{*}{500}  & Pretrained + ER  & 62.32 \scriptsize{$\pm$1.19} & 29.05 \scriptsize{$\pm$0.14} & 10.17 \scriptsize{$\pm$0.15} & 60.90 \scriptsize{$\pm$0.33}\\

& ER + TARC & \textbf{67.41} \scriptsize{$\pm$0.94} & \textbf{31.50} \scriptsize{$\pm$0.40} & \textbf{13.77} \scriptsize{$\pm$0.17} & \textbf{67.96} \scriptsize{$\pm$1.84} \\ \midrule

\multirow{2}{*}{5120}  & Pretrained + ER  &  \textbf{82.73} \scriptsize{$\pm$0.07} & \textbf{52.66} \scriptsize{$\pm$0.26}  & 27.98 \scriptsize{$\pm$0.20} & 72.18 \scriptsize{$\pm$7.85} \\

& ER + TARC & 82.21 \scriptsize{$\pm$0.38} & 52.30 \scriptsize{$\pm$0.20} & \textbf{32.04} \scriptsize{$\pm$0.37} & \textbf{75.28} \scriptsize{$\pm$1.06} \\ 

\bottomrule
\end{tabular}
\label{tab:pretraining}
\end{table*}

\section{Ablation study of robustness to natural corruptions}
\begin{figure}[h]
\centering
\includegraphics[width=0.4\linewidth, keepaspectratio]{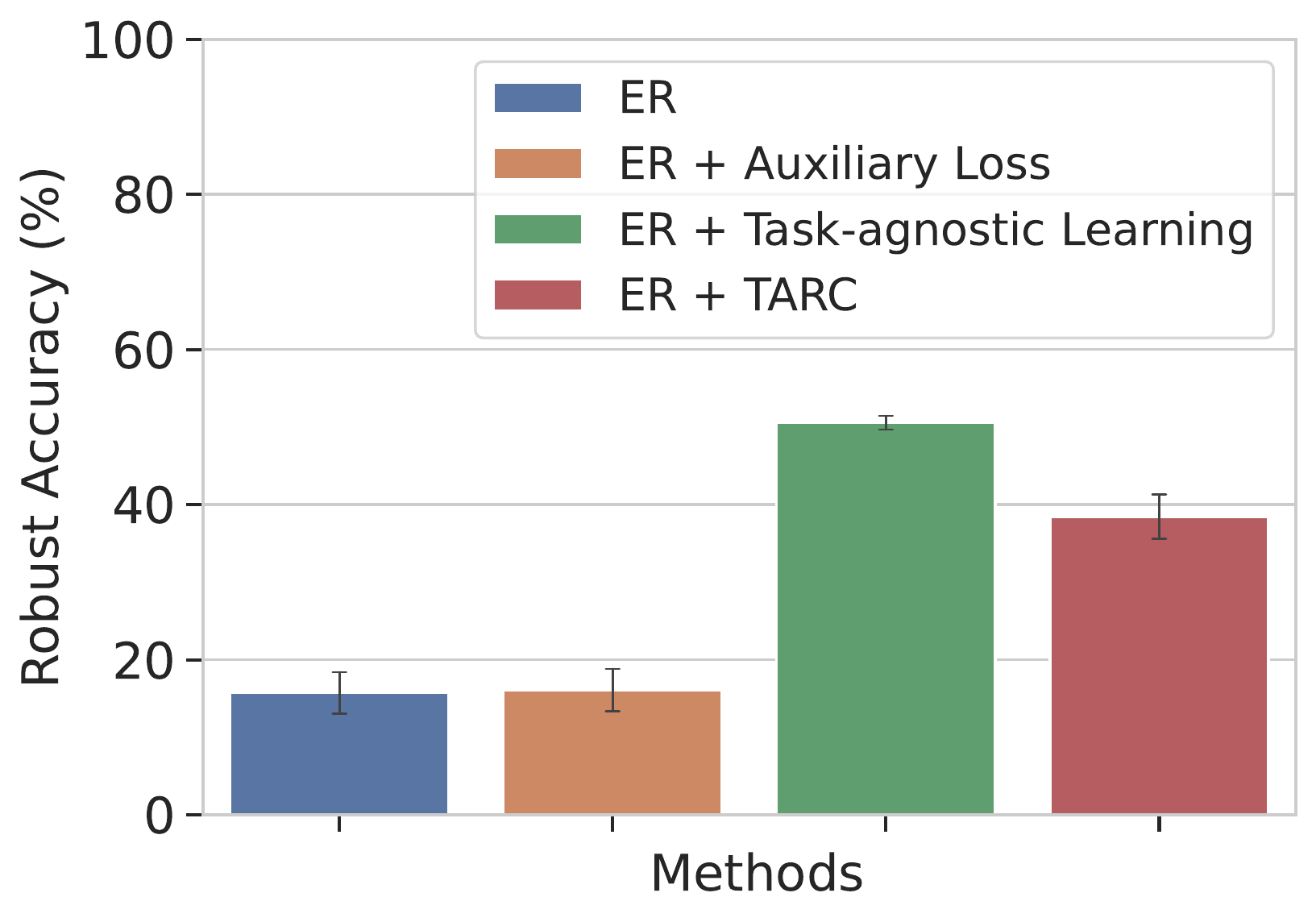}
\caption{Mean robust accuracy (\%) of CL models trained on CIFAR-10 with 500 buffer size against 19 natural corruptions.} \label{fig:ablation_robustness}
\end{figure}

We evaluate the individual components of our method against common image corruptions using CIFAR-10-C \citep{hendrycks2019}. Models are trained using clean CIFAR-10 with a buffer size of 500 and tested on CIFAR-10-C. We report the mean robust accuracy (\%) to evaluate the performance under natural corruptions \citep{hendrycks2019}. Figure \ref{fig:ablation_robustness} shows the mean robustness to 19 different corruptions averaged over five severity levels. ER with task-agnostic learning is considerably more robust to natural corruptions than with auxiliary task loss. However, as outlined in Section \ref{ablation_study_section}, ER with one of these components in isolation has a lower or marginally better natural accuracy than the baseline. On the other hand, the combination of these two components yields the best of both the worlds i.e. TARC achieves better natural accuracy while still being significantly robust to natural corruptions than ER baseline.

\section{Comparison of different auxiliary tasks}
\begin{figure}[h]
\centering
\includegraphics[width=0.6\linewidth, keepaspectratio]{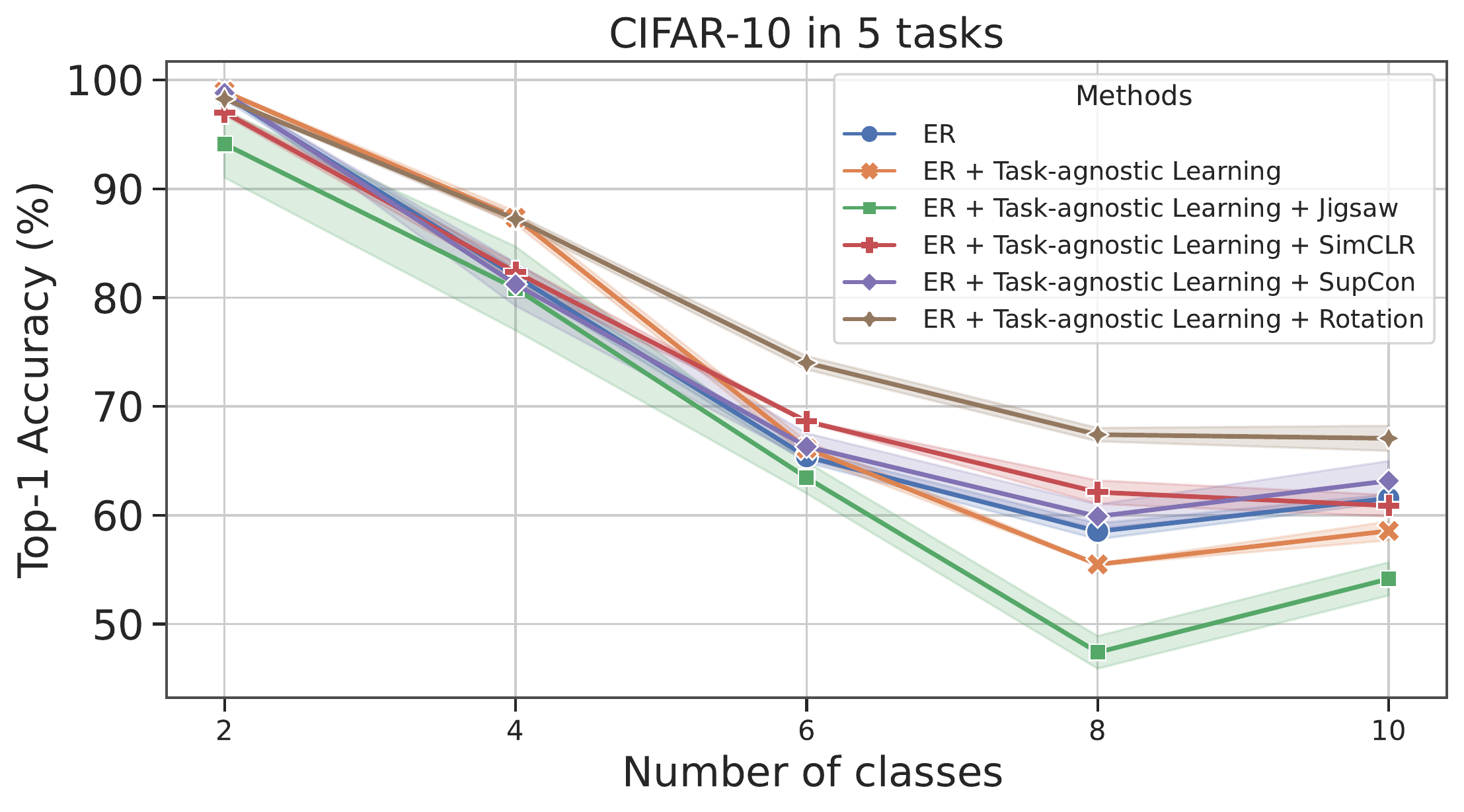}
\caption{Comparison of different auxiliary tasks in TARC framework. We also provide ER and ER + Task-agnostic learning baselines. Rotation prediction is best suited as an auxiliary task due to higher performance and low computational overhead.} \label{fig:ablation_auxiliary_loss}
\end{figure}

TARC framework consists of a task-specific learning phase wherein the CL model is trained using an auxiliary loss alongside cross-entropy objective. To better understand the intuition behind the choice of the auxiliary loss function, we compare two primitive and two sophisticated auxiliary losses in the TARC framework. Figure \ref{fig:ablation_auxiliary_loss} compares different CL models trained on CIFAR10 with buffer size 500 in the TARC framework. We also provide ER and ER+Task-agnostic Learning baselines for better comparison. As can be seen, Jigsaw prediction has the lowest performance while self-supervised algorithms such as SimCLR and SupCon have performances close to or slightly better than ER. Since CIFAR-10 images are small in size, jigsaw prediction turns out to be too difficult of a task to achieve.

On the other hand, rotation prediction helps preserve the generalizable features learned in the task-agnostic phase better, thereby leading to a significant improvement over ER baseline. The ability of rotation prediction to consolidate generalizable features has also been shown in other contemporary works (e.g. \citep{Zhu_2021_CVPR}).
We chose rotation prediction for its simplicity, limited computational overhead, and similarity to classification. In addition, rotation as an augmentation does not change the underlying semantics of the input and does not need any additional forward passes through the CL model.

\end{document}